\documentclass[conference]{IEEEtran}
\usepackage{cite}
\usepackage{amsmath,amssymb,amsfonts}
\usepackage{algorithmic}
\usepackage{graphicx}
\usepackage{textcomp}
\usepackage{xcolor}
\usepackage{fancyhdr}
\usepackage[hyphens]{url}
\usepackage{blindtext}
\usepackage{caption}
\usepackage{subcaption}

\def\BibTeX{{\rm B\kern-.05em{\sc i\kern-.025em b}\kern-.08em
    T\kern-.1667em\lower.7ex\hbox{E}\kern-.125emX}}

\newcommand{\hpcasubmissionnumber}{xxx}

\fancypagestyle{firstpage}{
  \fancyhf{}

  \fancyhead[C]{\normalsize{HPCA 2022 Submission
      \textbf{\#\hpcasubmissionnumber} \\ Confidential Draft: DO NOT DISTRIBUTE}} 
  \fancyfoot[C]{\thepage}
}

\title{Benchmarking Test-Time Unsupervised Deep Neural Network Adaptation on Edge Devices}

\author{
    \IEEEauthorblockN{Kshitij Bhardwaj, James Diffenderfer, Bhavya Kailkhura, and Maya Gokhale}
    \IEEEauthorblockA{Lawrence Livermore National Laboratory, Livermore, CA}
}

\begin{document}
\maketitle
\pagestyle{plain}


\begin{abstract}

The prediction accuracy of the deep neural networks (DNNs) after deployment at the edge can suffer with time due to {\em shifts} in the distribution of the new data. 
To improve robustness of DNNs, they must be able to update themselves to enhance their prediction accuracy. This adaptation at the resource-constrained edge is challenging as: (i) new labeled data may not be present; (ii) adaptation needs to be on device as connections to cloud may not be available; and (iii) the process must not only be fast but also memory- and energy-efficient. Recently, lightweight prediction-time unsupervised DNN adaptation techniques have been introduced that improve prediction accuracy of the models for noisy data by re-tuning the batch normalization (BN) parameters. This paper, for the first time, performs a comprehensive measurement study of such techniques to quantify their performance and energy on various edge devices as well as find bottlenecks and propose optimization opportunities. In particular, this study considers CIFAR-10-C image classification dataset with corruptions, three robust DNNs (ResNeXt, Wide-ResNet, ResNet-18), two BN adaptation algorithms (one that updates normalization statistics and the other that also optimizes transformation parameters), and three edge devices (FPGA, Raspberry-Pi, and Nvidia Xavier NX). We find that the approach that only updates the normalization parameters with Wide-ResNet, running on Xavier GPU, to be overall effective in terms of balancing multiple cost metrics. However, the adaptation overhead can still be significant (around 213 ms). The results  strongly  motivate the  need  for  algorithm-hardware co-design for efficient on-device DNN adaptation.\let\thefootnote\relax\footnotetext{This paper was selected for poster presentation in International Symposium on Performance Analysis of Systems and Software (ISPASS), 2022.}

\end{abstract}

\vspace{-0.1in}
\section{Introduction}
\label{sec:intro}
\vspace{-0.05in}



While deep neural networks (DNNs) are trained and validated extensively using a large dataset before deployment on edge, these networks are still prone to degradation in prediction accuracy in the post-deployment real world operation. The new input data samples that the trained DNNs encounter may have different distributions than their training dataset (called {\em dataset shifts})~\cite{bulusu2020anomalous}. This change in distributions could be caused due to sensor or environmental noise~\cite{wilson2020survey}. To improve robustness of DNNs, they are trained using data augmentation~\cite{hendrycks2019augmix} and adversarial training~\cite{kireev2021effectiveness} techniques. However, these offline robust training methods may not be enough to completely handle the real-time noise as they cannot cover the excessively wide range of potential data shifts that can occur post deployment.
Therefore, the neural networks need to be adapted to improve their prediction accuracy~\cite{wang2020tent, nado2020evaluating, diffenderfer2021winning}.

Various transfer learning techniques exist that can be used for neural network adaptation on edge devices, most of which however, require use of new labeled data that is seldom available at test time~\cite{sufian2021deep, zhang2021federated, meier2020transfer}. This paper targets challenging, albeit real-world scenarios where labels for the new, potentially domain-shifted, test data is not present: such as devices operating in remote places without human intervention~\cite{choi2020unsupervised, bhardwaj2021semi} or when the cost of annotating the new data with labels is too high and not feasible~\cite{he2021autoencoder, wilson2020survey}. Some key examples of these scenarios are: (i) DNNs performing human action recognition on drones without labeled samples~\cite{choi2020unsupervised}; (ii) techniques such as laser-induced breakdown spectroscopy in extreme environments (e.g., other planets)~\cite{bhardwaj2021semi}; and (iii) medical imaging where noise could be added due to scanners and the DNN for analysis needs to rapidly adapt without labeled data~\cite{he2021autoencoder}. 


In order to continually maintain or improve prediction accuracy of deployed DNNs and meet tight performance constraints of streaming applications, the DNNs must be adapted on device based on the new {\em shifted} test data. Due to hard deadlines, adaptation through cloud services may not be always feasible. Moreover, certain devices can be operating in areas with limited or no connectivity (e.g., military zones, other planets).
This prediction-time (or test-time) adaptation at the edge is challenging as: (i) new labeled data may not available; (ii) it should be fast as the networks are typically operating on streaming data with strict timing deadlines, and (iii) it should be lightweight and energy-efficient as the edge devices are generally resource-constrained and could be battery operated.

Recently, test-time unsupervised DNN adaptation techniques have been introduced that only update the batch-norm (BN) parameters of pre-trained DNNs based on newly seen data. BN layers are typically added to DNNs for faster and more robust training and are common in the modern DNNs. One such algorithm, which we call {\em BN-Norm}, recomputes the normalization statistics during test time~\cite{schneider2020improving, nado2020evaluating}. The other approach, which we call {\em BN-Opt,} not only re-estimates these statistics but also optimizes transformation BN parameters using a single backpropagation pass during prediction~\cite{wang2020tent}. Both of these algorithms have been shown to be effective in improving robustness for noisy data. The simplicity of these approaches make them ideal candidates for real-time adaptation at the edge. However, as these techniques are designed in isolation without considering/evaluating the hardware cost, it is unclear if they can run efficiently on edge devices.


{\bf Contributions.} In this paper, we perform a comprehensive measurement study of prediction-time unsupervised DNN adaptation techniques at the edge. To the best of our knowledge, this is the first study aimed at exploring such unsupervised approaches for resource-constrained devices. In  particular, we consider the following: (i) three  robust  DNNs that are pre-trained using data augmentation~\cite{hendrycks2019augmix} and adversarial training~\cite{kireev2021effectiveness} on CIFAR-10 image classification dataset (ResNeXt, Wide-ResNet, and ResNet-18). These models are among the top models on the robustbench leaderboard\footnote{Robustbench systematically tracks the progress being made in adversarial robust machine learning and keeps a record of the DNN models that perform the best in terms of prediction accuracy for corrupted/noisy datasets. However, it does not include adaptation approaches.}. 
They use float32 data type as robustness to corruptions has not been well explored for float16 or lower types; (ii) two  adaptation  algorithms  (BN-Norm~\cite{nado2020evaluating,schneider2020improving} and BN-Opt~\cite{wang2020tent}); (iii) three  edge  devices  with varying  compute  capability, memory, and price  (an Ultra96-v2 FPGA, a Raspberry-Pi, and Nvidia  Jetson  AI acceleration platform: Xavier  NX); and (iv) CIFAR-10-C dataset, that includes a variety of image corruptions, is used to test the models and the adaptation approaches.

The study answers the following algorithm-hardware co-design questions: (i) for each device, what is the optimal choice of robust DNN and test-time adaptation algorithm in terms of three objective functions: prediction accuracy, adaptation time, and energy dissipated during adaptation? (ii) what are the bottlenecks faced when executing these algorithms on the various devices? and (iii) are there potential ML and hardware optimization opportunities to improve adaptation time and energy? We also select the overall best DNN, adaptation technique, and device for varying constraints on the different cost metrics.
The results show that there exists a gap between algorithmic advances in adaptation and edge hardware designs to efficiently support such advancements. The study reveals interesting and somewhat non-obvious outcomes and demonstrates critical trade-offs between accuracy, performance, energy, and memory.

In particular, there are five key outcomes: {\bf (i)} for all three DNNs, BN-Opt outperforms BN-Norm in terms of reducing prediction errors during test time (by 2.45\% on average) for CIFAR-10-C. Both are significantly better than no adaptation (6.67\% and 4.02\%, respectively) even though the models were trained offline with robust methods; {\bf (ii)} while having a high number of BN parameters in a DNN (ResNeXt) leads to best prediction accuracy after adaptation, such models cannot be used with BN-Opt on memory-constrained edge devices such as Ultra96-v2 FPGA. Even when using the GPU on Xavier NX, ResNeXt and BN-Opt cause out of memory issues; {\bf (iii)} if for a given application, prediction accuracy is top priority, ResNeXt with BN-Opt should be used. For this configuration, Xavier NX leads to the lowest runtime while Raspberry Pi delivers the lowest energy; {\bf (iv)} if all three cost metrics are equally critical, then Wide-ResNet with BN-Norm should be used running on Xavier NX GPU, which outperforms other devices. Compared to (iii), this configuration is 220$\times$ faster and 114$\times$ more energy-efficient but with 5.6\% increase in prediction error. However, the extra adaptation time is still significant (213 ms) and can be a bottleneck for tight deadlines; {\bf (v)} embedded hardware DNN accelerators in edge devices can certainly improve performance and energy efficiency during prediction-time adaptation. The overhead for BN-Opt is more significant than BN-Norm, due to extra backpropagation, which becomes a considerable bottleneck when run on low-power Arm cores. Therefore, BN-Opt requires high-performance advanced accelerator, such as Volta GPU on Xavier, to accelerate this training step effectively, showing up to $7.89\times$ speedup compared to the CPU cluster on Xavier. The above outcomes strongly motivate the need for algorithm-hardware co-design when designing on-device DNN adaptation techniques. We conclude the paper with some architecture-algorithm insights gained from this study.


\section{Background}
\label{sec:back}

This section provides an overview of the techniques to improve robustness of DNNs as well as the BN-Norm and BN-Opt adaptation approaches that are the focus of this paper.

\vspace{-0.1in}
\subsection{Improving robustness of DNNs during training}
\label{subsec:robust}

There are two widely used offline training techniques to improve robustness of DNNs as described below.

\subsubsection{Data augmentation}
\label{subsub:augmix}

Data augmentation involves training a neural network using not only the ``clean'' original samples (such as images in CIFAR-10) but also with additional noisy versions of the original samples~\cite{hendrycks2019augmix}. This technique greatly improves the generalization performance of the DNNs. Common examples of augmentation include: {\em Cutout,} where a portion of an image is removed\cite{devries2017improved} and {\em Cutmix} that replaces a portion of an image with another~\cite{yun2019cutmix}. A recent augmentation approach that was shown to achieve better robustness is {\em Augmix}, which applies a series of augmentations to an image (such as rotate, posterize, etc.) then mixes the transformed images into a new image~\cite{hendrycks2019augmix}. Such samples, while quite different from the original, are not unrealistic and capture various noises. Augmix is used in this study. 

\subsubsection{Adversarial training}
\label{subsub:rlat}

This training makes a model robust to any adversarial attacks. The training involves solving a min-max optimization problem, where: (i) adversarial image samples are generated from the clean samples by finding imperceptible perturbations to the clean images such that the prediction loss of a DNN is maximized, and (ii) the prediction loss for these adversarial samples is minimized. In our study, we use a state-of-the-art adversarial training approach where imperceptibility is defined using learned perceptual image patch similarity (LPIPS) distance~\cite{kireev2021effectiveness}. LPIPS distance is based on the activations of a DNN evaluated on two different images and is more suitable to common image corruptions.


\vspace{-0.1in}
\subsection{BN-Norm test-time adaptation}
\label{subsec:norm}

BN-Norm adaptation is a lightweight approach to adapt DNNs at test time without supervision to improve their classification accuracy\cite{nado2020evaluating,schneider2020improving}. 
BN-Norm only updates the batch-norm (BN) layers without modifying the other layers. 
The BN statistics, mean ($\mu$) and standard deviation ($\sigma$), are typically fixed after training. However, in BN-Norm, these statistics are recomputed during test-time using an incoming batch of unlabeled data (other BN parameters are not modified). 

\vspace{-0.1in}
\subsection{BN-Opt test-time adaptation}
\label{subsec:tent}

BN-Opt also only updates the BN parameters during test-time without any supervision~\cite{wang2020tent}. 
BN-Opt, however, not only recomputes the mean and variance parameters for each of the batch-norm layers but also optimizes the {\em transformation parameters} that apply channel-wise scales and shifts to the features. This approach is efficient as the transformation parameters constitute $< 1\%$ of the total model parameters.

In particular, two steps are performed for each batch-norm layer during test-time: (i) normalization that standardizes the input $x$ into $x' = (x-\mu)/\sigma$ using its mean and standard deviation, and (ii) transformation that turns $x'$ into $x'' = \gamma x' + \beta$ using scale ($\gamma$) and shift ($\beta$) parameters. While the statistics for (i) are recomputed from the unlabeled data, scale and shift parameters in (ii) are optimized by a loss function. Since, the optimization is performed using unlabeled data, {\em entropy of model predictions} is used as the loss function. Shannon entropy for a prediction $y$, which can be computed without any labeled data, is defined as: $H(y) = - \sum_{c} p(y_c)log p(y_c)$ for probability of $y$ for class $c$. Predictions with lower entropy are shown to exhibit lower prediction error rate as well. 

\vspace{-0.1in}
\section{Measuring DNN Adaptation at the Edge}
\label{sec:dse}

This section provides an overview of our measurement study, followed by the details on the various parameters.

\subsection{Measurement study overview}
\label{subsec:overview}

\begin{figure}[t]
\centering
  \includegraphics[width=1\columnwidth]{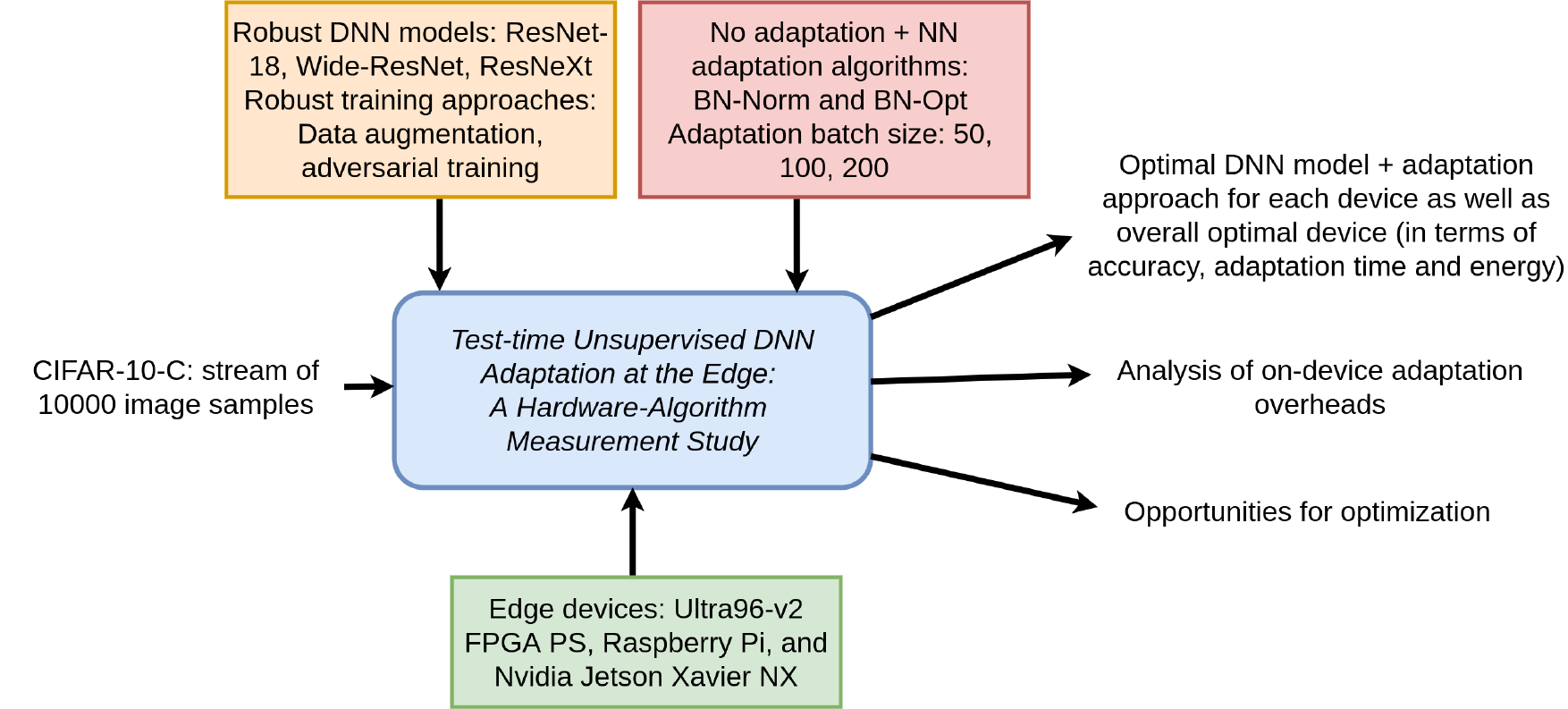}
  \caption{Measurement study overview\vspace{-0.2in}} 
  \label{fig:overview}
\end{figure}

Figure~\ref{fig:overview} shows an overview of our hardware-algorithm measurement study for test-time unsupervised DNN adaptation at the edge. The aim is threefold: (i) for each target device, among the state-of-the-art robust DNN models and adaptation strategies, find the best model and adaptation algorithm that leads to low prediction errors as well as low adaptation runtime and energy consumption for a given noisy data stream. While the former cost metric is important for robustness, minimizing the latter two are critical to meet the constrained timing requirements for various streaming applications and be able to run on low-power devices. We also find the overall best DNN, adaptation scheme, and device in terms of the three cost metrics; (ii) measure and provide a detailed analysis of the adaptation time overhead for the different models and algorithms running on the various edge devices; and (iii) find algorithm and hardware optimization opportunities.

In this paper, we consider three state-of-the-art robust neural networks, two unsupervised NN adaptation approaches, a well-known image classification dataset with corruptions, and three edge devices. The robust DNNs are pre-trained models: trained with techniques such as Augmix data augmentation or adversarial training. 
The two adaptation approaches are: BN-Norm adaptation (Section~\ref{subsec:norm}) and BN-Opt adaptation (Section~\ref{subsec:tent}). In addition, we also include no adaptation approach to provide a baseline for comparisons. For each adaptation algorithm, the amount of recently-seen unlabeled test data used for model update (batch size) is also varied: 50, 100, and 200 samples. We use CIFAR-10-C dataset~\cite{croce2020robustbench} that applies 15 different common corruptions to the standard CIFAR-10 image classification dataset. Finally, the three edge devices considered have varying processing capability and memory (as well as power requirements), and are typically deployed for a variety of streaming applications. These devices are: (i) with no inherent DNN acceleration capability (Ultra96-v2 FPGA and Raspberry-Pi), and (ii) with a DNN accelerator (Nvidia Jetson Xavier NX). 

\subsection{{Robust} DNN models}
\label{subsec:dse_robust}

\subsubsection{ResNet-18}
\label{subsubsec:r18}
To improve robustness, ResNet-18 is trained using both Augmix data augmentation technique for CIFAR-10 dataset (Section~\ref{subsub:augmix}, similar to~\cite{hendrycks2019augmix}) as well as using LPIPS-based adversarial training (Section~\ref{subsub:rlat}, similar to~\cite{kireev2021effectiveness}). This trained ResNet-18 has 0.56 Giga multiply-accumulate (GMAC) operations and 11.17M total parameters, out of which there are 7808 batch-norm parameters (target for BN-Norm and BN-Opt). Overall it takes 86 MB memory.

\subsubsection{Wide-ResNet}
\label{subsubsec:wrn}
We leverage Wide-ResNet-40-2 model, trained using 
Augmix data augmentation technique for CIFAR-10 (Section~\ref{subsub:augmix}, similar to~\cite{hendrycks2019augmix}). This model has 0.33 GMAC operations, 2.24M total parameters, 5408 batch-norm parameters, and takes 9 MB memory.

\subsubsection{ResNeXt}
\label{subsubsec:rxt}
We use ResNeXt29\_32x4d 
with a cardinality of 4 and a base width of 32. 
ResNeXt is also trained using Augmix for CIFAR-10 (Section~\ref{subsub:augmix}, similar to~\cite{hendrycks2019augmix}). This model has 1.08 GMAC operations, 6.81M total parameters, 25216 batch-norm parameters, and takes 26 MB memory.
ResNeXt includes significantly more batch-norm parameters than the above two models.

\subsection{Image classification data with corruptions}
\label{subsec:dse_cifar10c}

While the above three models are pre-trained on CIFAR-10, the CIFAR-10-C dataset is used for testing the models. CIFAR-10-C includes 15 different types of corruptions, such as gaussian noise, blur, snow, fog, frost, brightness, etc., with 5 different severity levels (5 being the most severe and 1 being the least). In our study, all 15 corruptions (level 5) are used. 

The test data consist of streaming 10000 unlabeled CIFAR-10-C image samples for each corruption type. To perform online adaptation during test time, batches of recently seen image samples are used by the adaptation algorithms. Three different batch sizes are used which have a direct impact on the adaptation efficacy as well as performance/energy/memory during adaptation: 50, 100, and 200. 


\subsection{Adaptation techniques}
\label{subsec:dse_adapt}

In this study, a no adaptation and two DNN adaptation algorithms are used on the CIFAR-10-C test data: (i) no adaptation, where the three pre-trained robust DNNs are used without any changes; (ii) BN-Norm adaptation (Section~\ref{subsec:norm}) that uses the recently seen batch of image samples to simply recompute the mean and standard deviation BN parameters; and (iii) BN-Opt adaptation (Section~\ref{subsec:tent}) which recalculates the normalization parameters as well as optimizes the BN transformation parameters. While BN-Norm does not need to retrain when adapting the model, BN-Opt, on the other hand, performs a {\em single backpropagation pass} using the collected batch of image samples to optimize the transformation parameters for the entropy loss function using the Adam optimizer. 

The algorithms are implemented using Pytorch~\cite{torch}. During test time with CIFAR-10-C, the no adaptation algorithm keeps the model in {\em eval()} mode of Pytorch (inference only), while BN-Norm and BN-Opt require the model to be in {\em train()} mode (to allow for adaptation with inference). The {\em forward} passes for both BN-Norm and BN-Opt are modified to adapt the DNN: at each adaptation point, these algorithms first perform inference followed by updating (and optimizing in case of BN-Opt) the batch-norm parameters based on the collected recently seen batch of data. Since our focus is on adaptation (or retraining) on edge devices, optimized frameworks for fast inference on edge, such as Tensorflow-Lite~\cite{tflite} and TensorRT~\cite{tensorrt}, are not applicable and hence not used.

\subsection{Edge devices}
\label{subsec:dse_edge}

Details on the three edge devices are presented next.

%

\subsubsection{Ultra96-v2 FPGA}
\label{subsubsec:fpga}

We use an FPGA due to their increased use in streaming application~\cite{wu2019fpga}, both due to the support for programmability as well as low-power operations. The Ultra96-v2 consists of a Xilinx Zynq UltraScale+ MPSoC ZU3EG A484 processing system (PS) part and 2 GB LPDDR4 memory. The DNN adaptation algorithms are run on the PS while the programming logic (PL) part is not used in this study. 
We used the PS quad-core ARM Cortex A53 multi-processing system that can run up to 1.5 GHz, with 32 KB L1 caches and a 1 MB L2 cache. On the software side, we use the {\em Pynq} framework~\cite{pynq} with  {\em Petalinux} to create the Linux image on the PS. Pytorch-1.8.0 for ARM is used.

\subsubsection{Raspberry-Pi}
\label{subsubsec:rpi}

Another low-power edge device that has become critical for streaming applications is Raspberry-Pi~\cite{hu2020fast}. We use the Raspberry-Pi 4 Model B that consist of a quad-core ARM A72 SoC running at up to 1.5 GHz with 32 KB data + 48 KB instruction L1 caches and 1 MB L2 cache. This device also comes with 8 GB LPDDR4 memory. We booted Ubuntu 21.04 and installed Pytorch-1.8.0 for ARM.




\subsubsection{Nvidia Jetson Xavier NX}
\label{subsubsec:nx}

While the above two devices did not use hardware neural network acceleration capability, NX includes a GPU to accelerate DNN processing.
NX uses a 6-core Nvidia CARMel ARM SoC (128 KB L1 instruction cache + 64 KB L1 data cache, and 6 MB L2 cache and 4 MB L3 cache) that can run up to 1.9 GHz speed. It integrates a 384-core Volta GPU (maximum frequency of 1.1GHz) and 8GB LPDDR4 memory. NX uses Jetpack 4.4 software development kit (SDK) that boots Linux4Tegra operating system (similar to Ubuntu 18.04 but for Nvidia devices) and comes with CUDA 10.2 and GPU NN acceleration libraries (cuDNN 8.0). Pytorch-1.8.0 is installed using Nvidia's pre-built installer. 

\subsection{Objective functions}
\label{subsec:objectives}

Three objective functions are considered: overall prediction accuracy for the entire test stream, average forward time per batch (inference + any adaptation), and average energy consumption per batch. Accuracy and performance are computed using Pytorch. To compute energy, we measure power per batch using Kuman wall outlet power meter.

Different applications give priorities to different objectives. For example, some scenarios may have strict performance constraints, hence adaptation times should be minimized, while for others maintaining prediction accuracy during noise is more important. To find the best DNN model and adaptation approach for a variety of cases, we combine the multiple objective functions into a single one using a weighted approach. In particular, we use: $w_1*time + w_2*energy + w_3*pred\_error$ as the single objective function to minimize, where $w_1$, $w_2$, and $w_3$ are weights whose sum should be equal to 1. We find optimal configurations for four cases that cover a wide variety of scenarios: (i) same importance to all three objectives (each weight equal to 0.33); (ii) performance is the top priority but other metrics are not completely discarded and hence assigned much smaller weights ($w_1=0.8$ and others are 0.1 each); (iii) prediction error is more important ($w_3$ is 0.8) with other metrics much less critical but still considered (0.1 each); and (iv) energy is critical with 0.8 weight and others as 0.1. 
\vspace{-0.1in}
\section{Measurement Results and Analysis}
\label{sec:results}
\vspace{-0.05in}

This section first presents the prediction accuracy results for the different robust DNNs and adaptation approaches, followed by their detailed performance analysis and multi-objective cost trade-offs for each of the three edge devices. The section concludes with a summary of the overall outcomes of our study, a comparison with standard MobileNet-V2 model, and the architecture-algorithm insights gained from this study.

\vspace{-0.1in}
\subsection{Prediction accuracy}
\label{subsec:err}

\begin{figure}[t]
\centering
  \includegraphics[width=0.8\columnwidth,trim=4 4 4 4,clip]{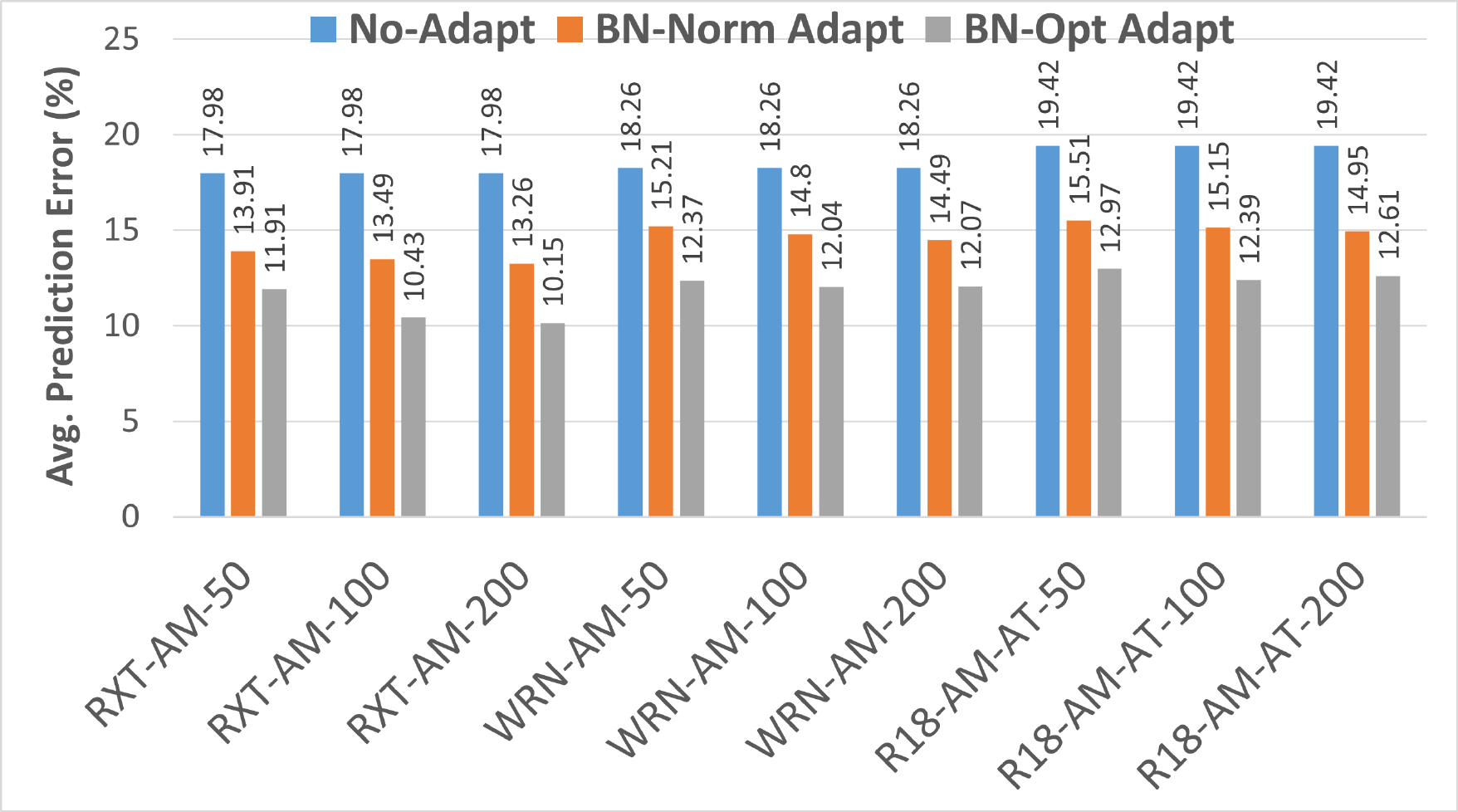}
  \caption{Average prediction errors for CIFAR-10-C\vspace{-0.2in}}
  \label{fig:pred_err}
\end{figure}

Figure~\ref{fig:pred_err} shows the prediction error for the three algorithms: no adaptation, BN-Norm, and BN-Opt, for an entire stream of 10000 unlabeled CIFAR-10-C image samples. The error is averaged across the 15 noise types. Three robust NNs are considered: ResNeXt with AugMix~\cite{hendrycks2019augmix} (RXT-AM), Wide-ResNet with AugMix~\cite{hendrycks2019augmix} (WRN-AM), and ResNet-18 with both AugMix and adversarial training~\cite{kireev2021effectiveness} (R18-AM-AT). For each network, three test-time batch sizes are used for online adaptation: 50, 100, and 200. The average prediction error is found to be consistent across the edge devices in cases where they do not run out of memory (RPi, Xavier NX).

As evident, no adaptation algorithm, even when using AugMix and adversarial training techniques, shows the worst prediction error. BN-Norm, which recomputes the normalization parameters during test time, is able to improve on no adaptation by an average of 4.02\% (across the 9 cases). BN-Opt, on the other hand, shows an average improvement of 6.67\% over no adaptation and 2.65\% over BN-Norm as it also optimizes the transformation parameters. For both BN-Norm and BN-Opt, reduction in prediction error going from 50 to 100 batch size is higher than from 100 to 200 showing that larger batch sizes than 200 will yield diminishing returns. Additionally, bigger batch sizes for online adaptation may not be reasonable for memory-constrained devices. 


\subsection{Ultra96-v2 processing system (PS)}
\label{subsec:pynq}

\begin{figure}[t]
\centering
  \includegraphics[width=1\columnwidth,trim=4 4 4 4,clip]{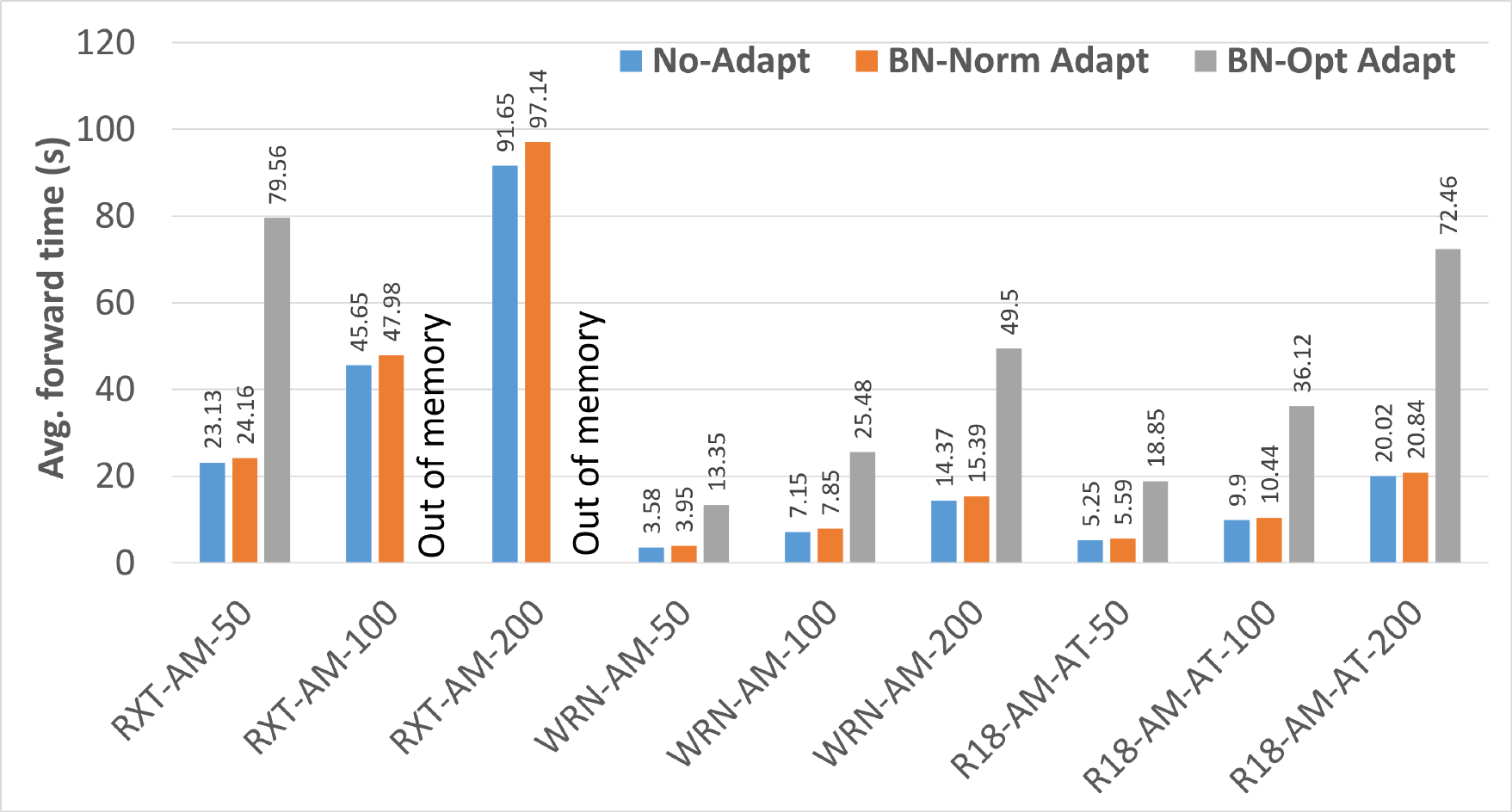}\vspace{-0.15in}
  \caption{Ultra96 PS forward times (inference + any adaptation)\vspace{-0.25in}} 
  \label{fig:fpga}
\end{figure}

This section presents the performance analysis for the FPGA PS, followed by the performance-energy-accuracy trade-offs discussion. Average forward time per batch is used as the performance metric to compare the different neural networks using no adaptation (No-Adapt), and BN-Norm and BN-Opt. Forward time for No-Adapt is simply the inference latency while for the others it also includes any extra adaptation time: re-estimating normalization parameters for BN-Norm and BN-Opt, and a single backpropagation pass for BN-Opt to optimize the transformation parameters. The times are averaged over multiple adaptations performed over an input data stream of CIFAR-10-C samples. Pytorch is run on the FPGA PS in a multi-threaded setting utilizing all four Arm A53 cores, while exploiting intra-operation parallelism. Similarly, energy is also measured for No-Adapt (inference only) and BN-Norm and BN-Opt (inference+adaptation) at the different adaptation points and then averaged.

{\bf Performance analysis.} As shown in Figure~\ref{fig:fpga}, compared to No-Adapt, the extra average adaptation overhead for BN-Norm is 1.40 secs 
while for BN-Opt it is much higher: 30.27 secs. 
The simple recompute operation of the former makes it more suitable for test-time NN adaptation on an A53 device (while incurring a prediction error overhead of around 2.65\%). 
In addition, while BN-Norm is able to run for all 9 cases on the FPGA PS, BN-Opt however, runs out of memory for RXT for 100 and 200 batch sizes. So, while RXT-AM-100/200 with BN-Opt achieved the lowest prediction error (Figure~\ref{fig:pred_err}), they cannot be used on a constrained memory device with only 2 GB memory. Even though RXT's model size is 26 MB, lower than R18 (86 MB), it runs out of memory as the number of BN parameters optimized through backpropagation for RXT are significantly higher than the others (25216 vs. 5408/7808). To perform backpropagation, Pytorch creates a dynamic computational graph during forward pass that includes the nodes that are enabled for gradient computation. RXT's graph takes 3.12 GB and 5.1 GB memory (based on a memory profiler used for R-Pi and NX) with batch size of 100 and 200, respectively, which is more than the FPGA's 2GB memory.
Moreover, compared to the other ResNet models, RXT also shows significantly higher forward time due to many more number of MAC operations (RXT: 1.08 GMACs, WRN: 0.33 GMACs, R18: 0.56 GMACs).

\begin{figure}[t]
\centering
  \includegraphics[width=0.8\columnwidth,trim=4 4 4 4,clip]{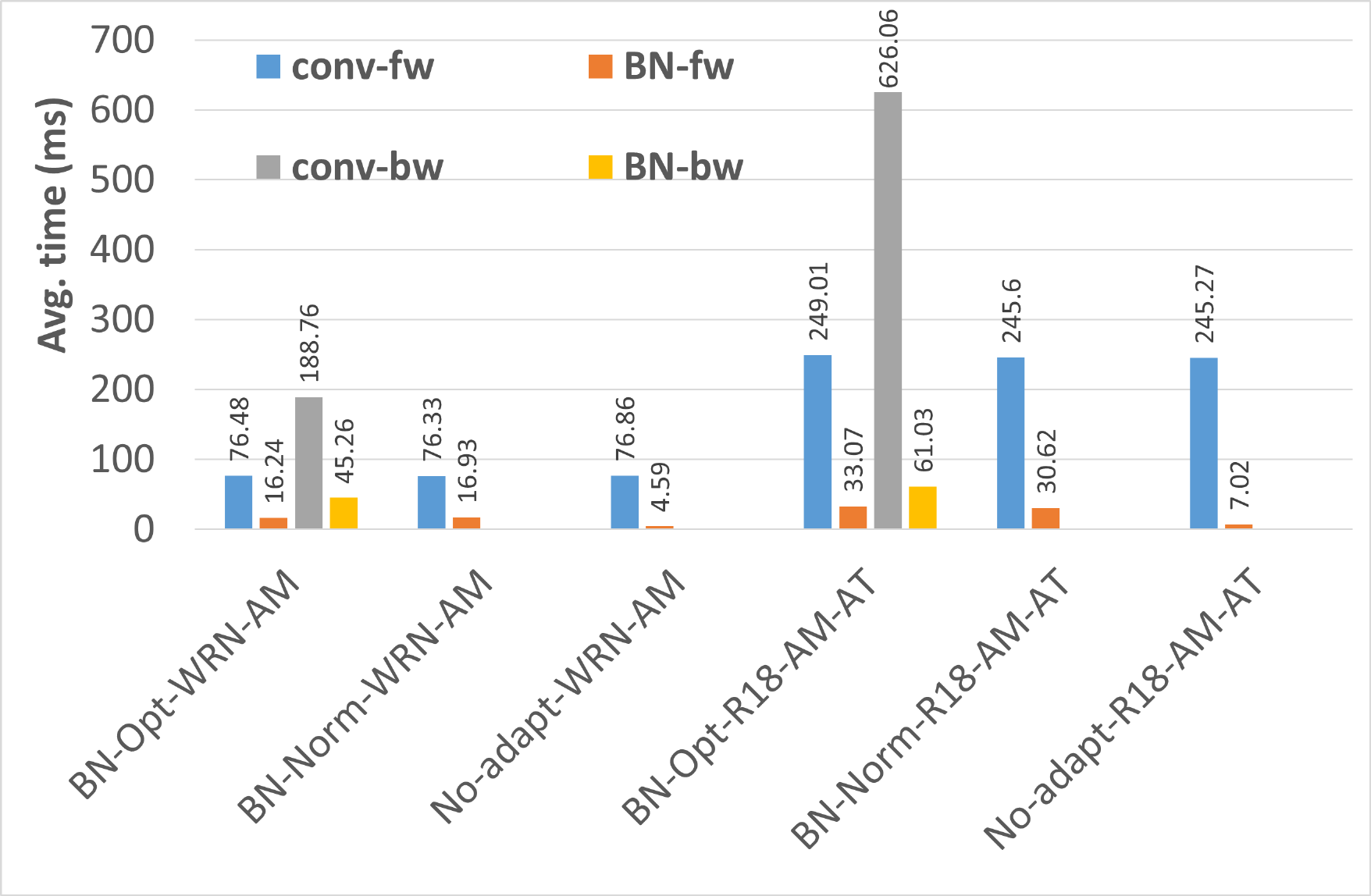}\vspace{-0.15in}
  \caption{Forward (fw) and any backpropagation backward (bw) pass performance on Ultra96-v2 for batch size 50\vspace{-0.2in}}
  \label{fig:pynq_times}
\end{figure}

Figure~\ref{fig:pynq_times} shows the breakdown of performance for BN-Opt, BN-Norm, and No-Adapt for Wide-ResNet and ResNet-18 in terms of the time spent on forward and backpropagation backward passes (averaged for convolution and BN layers). This breakdown is obtained using the Pytorch Autograd profiler with batch size 50 (similar trend is expected for other batch sizes). The profiler runs out of memory for RXT-AM (hence not included). While the forward times for convolution layers are almost the same for a given network for each of the three algorithms, the BN forward time is higher for BN-Norm and BN-Opt (up to 3.68$\times$ for WRN, 4.71$\times$ for R18) compared to no adaptation. The reason is the re-tuning of the normalization parameters in the forward pass which adds a significant cost (higher for R18 as it has more number of BN parameters). Moreover, while there is no backpropagation performed for No-Adapt and BN-Norm, this backward pass of BN-Opt shows considerable overhead for both convolution and BN layers compared to the forward passes (up to 2.51$\times$ and 2.78$\times$, respectively). Although this result is expected as A53s are not designed to run compute-intensive training algorithms, it also motivates the need to explore potentially offloading training kernels to the PL side of the FPGA for acceleration. Some of the recent works have started looking into training acceleration on FPGAs~\cite{luo2019towards}. 

\begin{figure}[t]
\centering
  \includegraphics[width=1\columnwidth,trim=4 4 4 4,clip]{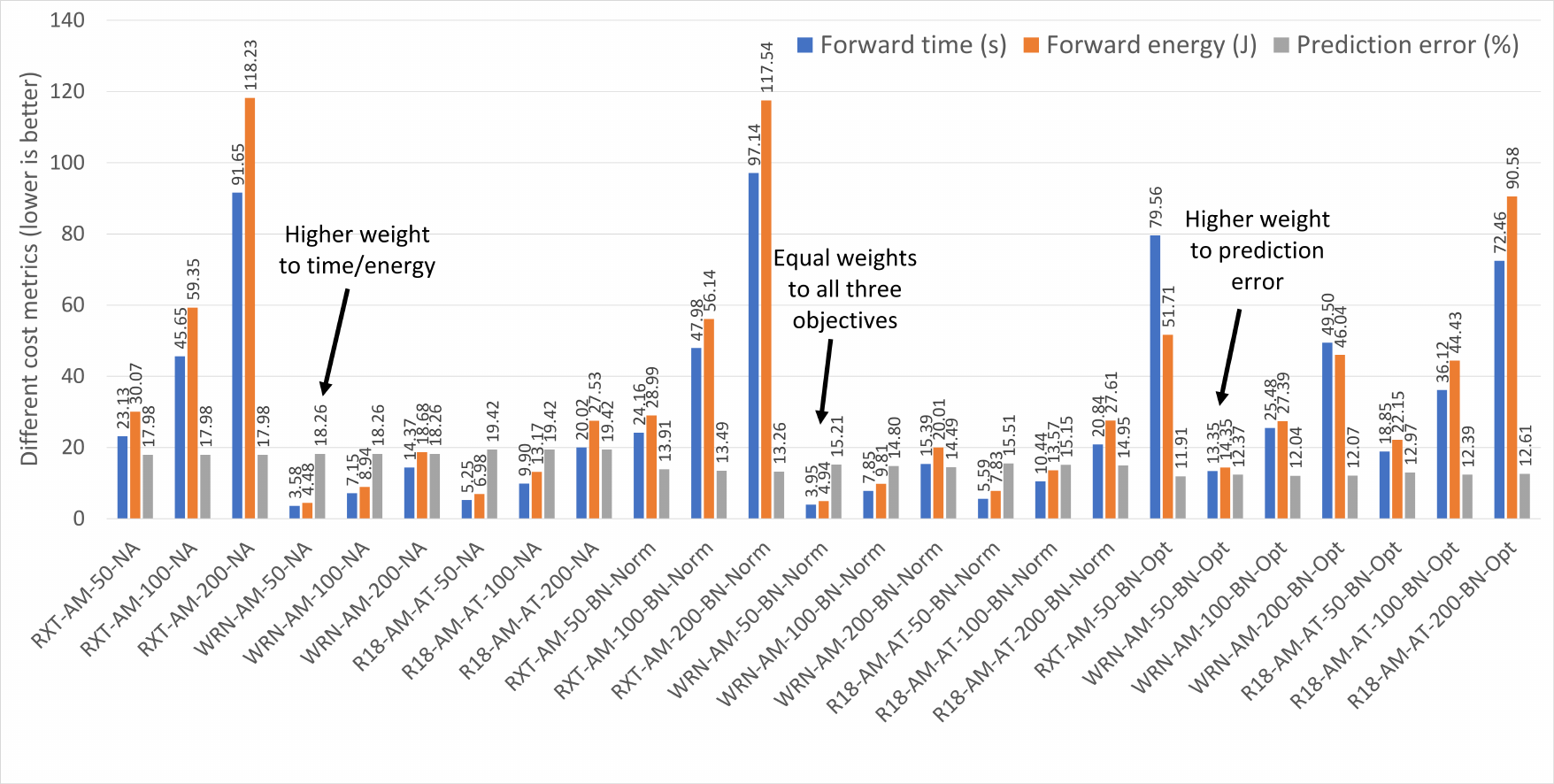}
  \vspace{-0.4in}
  \caption{Performance-energy-accuracy trade-offs: Ultra96-v2 PS\vspace{-0.2in}}
  \label{fig:pynq_tradeoffs}
\end{figure}

{\bf Performance-energy-accuracy trade-offs.} Figure~\ref{fig:pynq_tradeoffs} shows all three objectives for the various cases of adaptation algorithms, DNNs, and batch sizes. Note that RXT-AM-100/200 for BN-Opt are not present as they ran out of memory. Three important cases are shown based on the single weighted objective function ($w_1*forward\_time + w_2*energy + w_3*error$, Section~\ref{subsec:objectives}): (i) all three objectives are assigned a weight of 0.33 each, which leads to WRN-AM-50 with BN-Norm to be the optimal point (3.95 secs, 4.93 J, and 15.21\% error) because of lightweight adaptation of BN-Norm, coupled with the smallest number of BN parameters in Wide-ResNet; (ii) when accuracy is assigned a higher weight of 0.8 while still giving small weights of 0.1 each to the other two, WRN-AM-50 with BN-Opt is selected (13.35 secs, 14.35 J, 12.37\% error) as BN-Opt is highly effective at reducing prediction error and WRN balances accuracy with performance and energy better than other models; 
and (iii) when either performance or energy are top priority (0.8 weight), WRN-AM-50 with No-Adapt (3.58 secs, 4.47 J, 18.26\% error) is the reasonable choice. 

{\bf Summary.} Even for Ultra96-v2 with A53 cores and 2 GB memory, real-time unsupervised adaptation is possible although not efficient. Wide-ResNet using BN-Opt (batch size of 50) is optimal when priority is given to prediction error. In contrast, BN-Norm for the same network is the best when equal weight is given to prediction error, performance, and energy. While ResNeXt (highest number of BN parameters) with BN-Opt achieved the lowest prediction error, it is unable to run on Ultra96-v2 PS, showing the need for device-aware design of robust DNN models and adaptation algorithms targeting memory-constrained devices. BN-Opt incurs significant performance overheads over no adaptation and BN-Norm as it involves backpropagation. Use of PL side of the FPGA to offload training kernels can be explored.

\subsection{Raspberry Pi (RPi) 4}
\label{subsec:rpi}

Performance analysis and the various cost trade-offs are now presented for RPi.

\begin{figure}[t]
\centering
  \includegraphics[width=1\columnwidth,trim=4 4 4 4,clip]{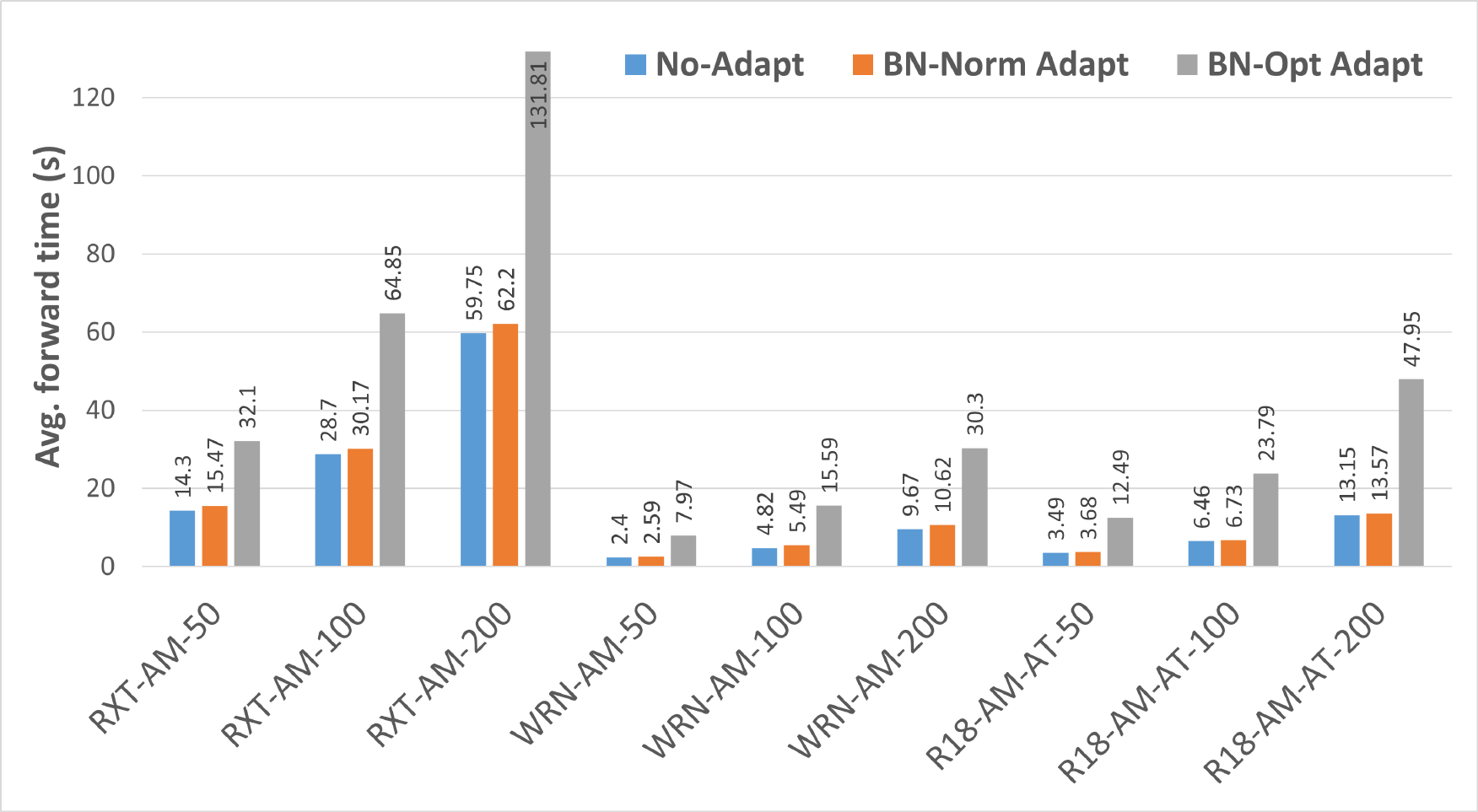}\vspace{-0.15in}
  \caption{RPi forward times (inference + any adaptation time)\vspace{-0.2in}}
  \label{fig:rpi_adapt}
\end{figure}

{\bf Performance analysis.} As shown in Figure~\ref{fig:rpi_adapt}, all three DNNs, with both BN-Norm and BN-Opt, are able to run on the RPi as it packs more memory (8GB) than the FPGA. Pytorch is run multi-threaded on quad-core Arm A72 cores. Compared to No-Adapt, the extra adaptation time (in addition to the inference, across all 9 cases) for BN-Norm is only 0.86 secs on average 
but for BN-Opt it is 24.9 secs. 
Due to the use of A72s, these times are reduced compared to the FPGA. 

\begin{figure}[t]
\centering
  \includegraphics[width=1\columnwidth,trim=4 4 4 4,clip]{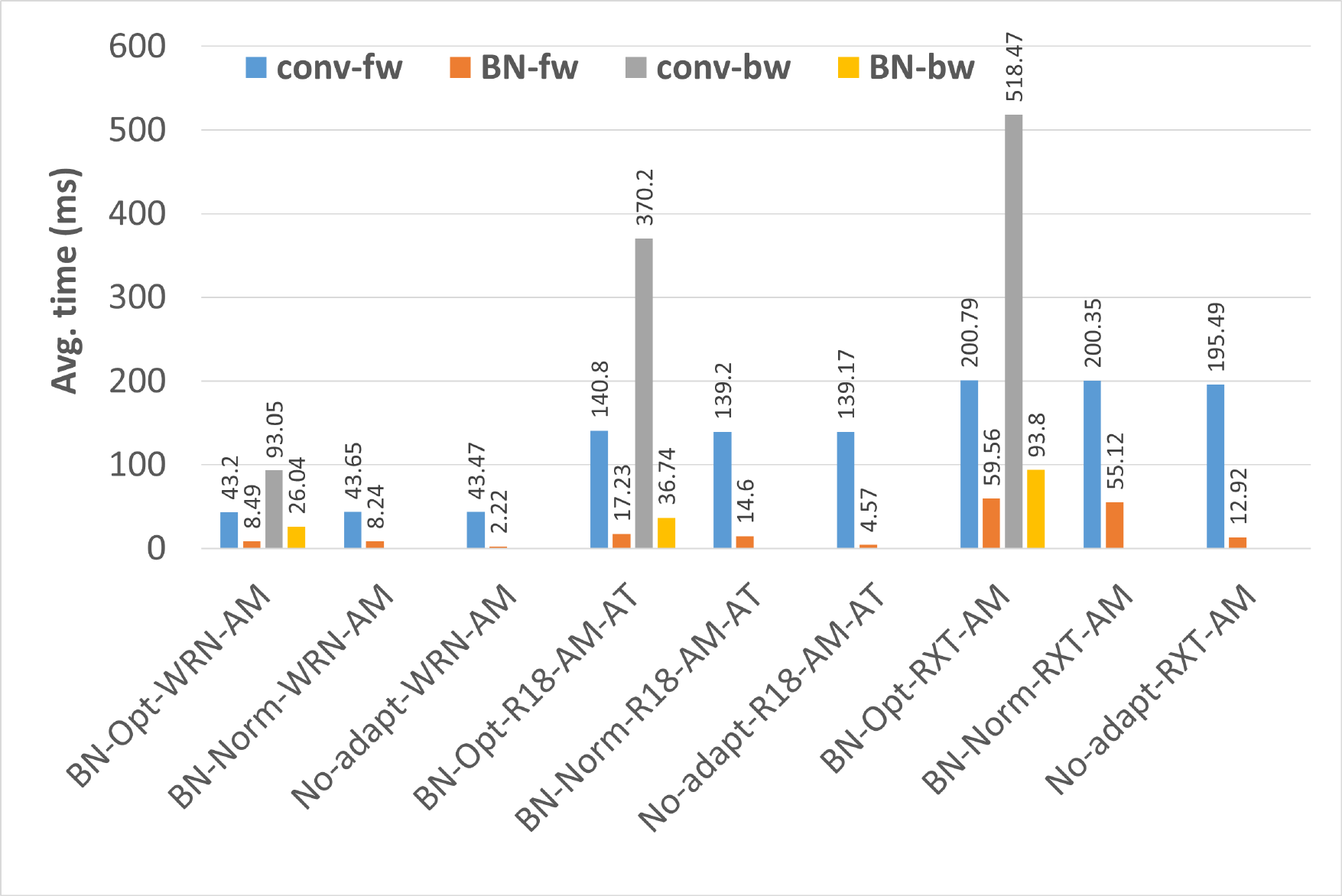}
  \vspace{-0.3in}
  \caption{Forward (fw) and any backpropagation backward (bw) pass performance for RPi for batch size 50\vspace{-0.2in}}
  \label{fig:rpi_times}
\end{figure}

Figure~\ref{fig:rpi_times}  shows  the  breakdown  of  performance  for BN-Opt, BN-Norm,  and  No-Adapt approaches for all three models in terms of the average time spent on forward and backpropagation backward passes. Pytorch Autograd profiler is used with the batch size fixed to 50 (similar trends for other batch sizes). Similar to Ultra96-v2, forward BN time for BN-Norm and BN-Opt is up to 4.6$\times$ higher than No-Adapt. Backward times for convolutions and BN in BN-Opt (zero for BN-Norm and No-Adapt) are also significant and explain the high adaptation times for BN-Opt in Figure~\ref{fig:rpi_adapt}. 

\begin{figure}[t]
\centering
  \includegraphics[width=1\columnwidth,trim=4 4 4 4,clip]{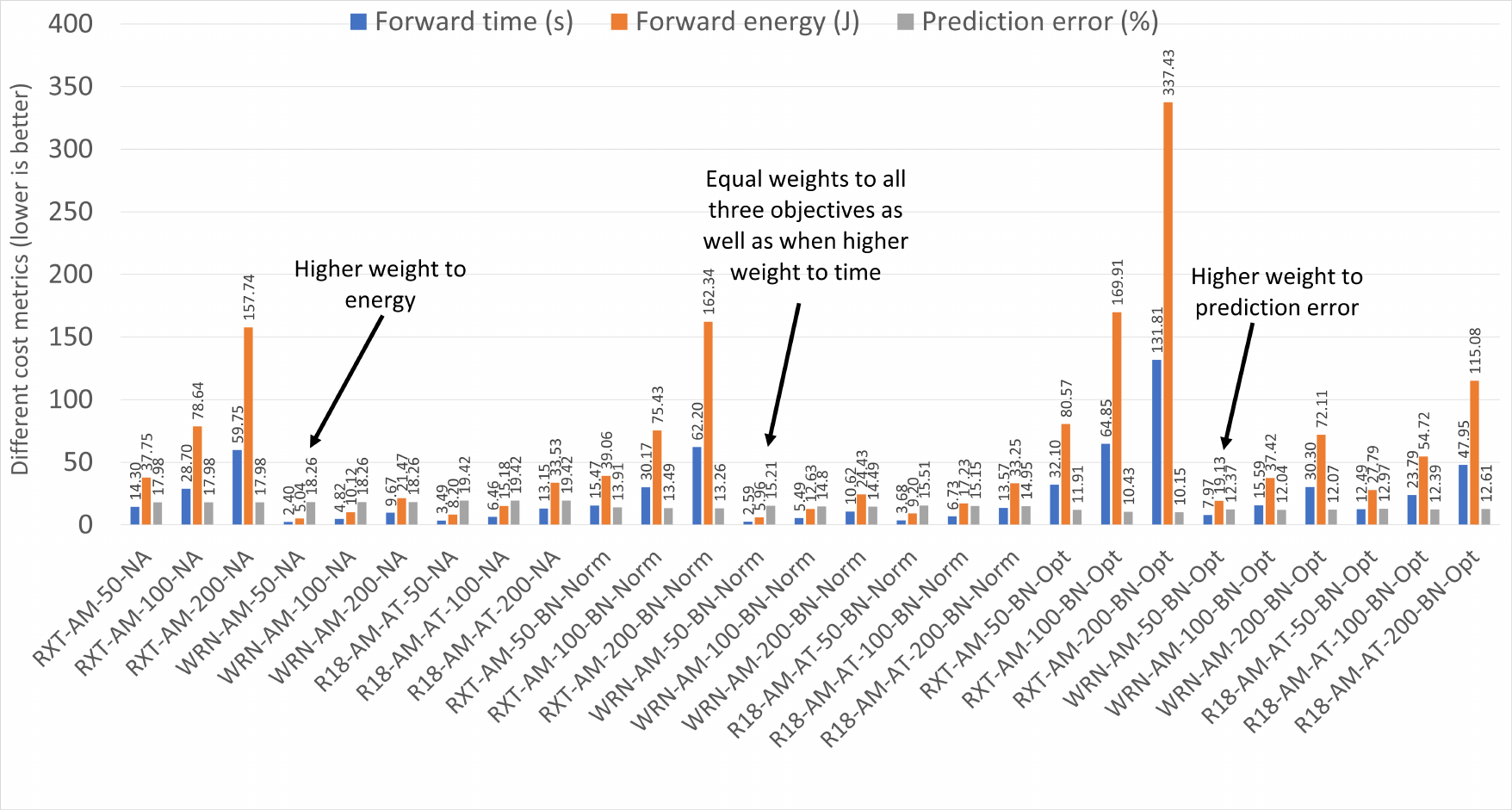}
  \vspace{-0.3in}
  \caption{Performance-energy-accuracy trade-offs analysis: RPi\vspace{-0.0in}}
  \label{fig:rpi_tradeoffs}
\end{figure}

{\bf Performance-energy-accuracy trade-offs.} Figure~\ref{fig:rpi_tradeoffs} shows all three objectives for the various cases. The same weighted multi-objective optimization function is used. The important outcomes are: (i) for all three objectives assigned an equal weight of 0.33 each, WRN-AM-50 with BN-Norm is the best (2.59 secs, 5.95 J, and 15.21\% error) because of lightweight WRN and fast adaptation of BN-Norm; (ii) when accuracy is assigned a higher weight of 0.8 while the other two are 0.1 each, WRN-AM-50 with BN-Opt is selected (7.97 secs, 19.12 J, 12.37\% error) as BN-Opt is highly effective at reducing prediction error for noisy data. However, this configuration shows 3.07$\times$ higher forward time and 3.21$\times$ more energy than (i);
(iii) when performance is the top priority (using 0.8 weight), interestingly, WRN-AM-50 with Norm is again selected. Although its forward time is very slightly higher than No-Adapt for WRN-AM-50 (2.59 secs vs. 2.04 secs) but due to the 0.1 weight assigned to accuracy, BN-Norm leads to the overall better configuration; and (iv) when energy is assigned the highest weight (0.8), WRN-AM-50 with no adaptation is the best choice (2.4 secs, 5.04 J, 18.26\%). 

{\bf Summary.} Bigger memory in RPi allows models with higher number of BN parameters, such as ResNeXt, to run with BN-Opt. 
BN-Norm, with small loss of accuracy, can be used to perform online adaptation more efficiently than BN-Opt. Therefore, while Wide-ResNet with BN-Opt is the best choice when the top priority is accuracy, the same network with BN-Norm should be used when fast adaptation is key. 

\begin{figure}[t]
\centering
  \includegraphics[width=1\columnwidth,trim=4 4 4 4,clip]{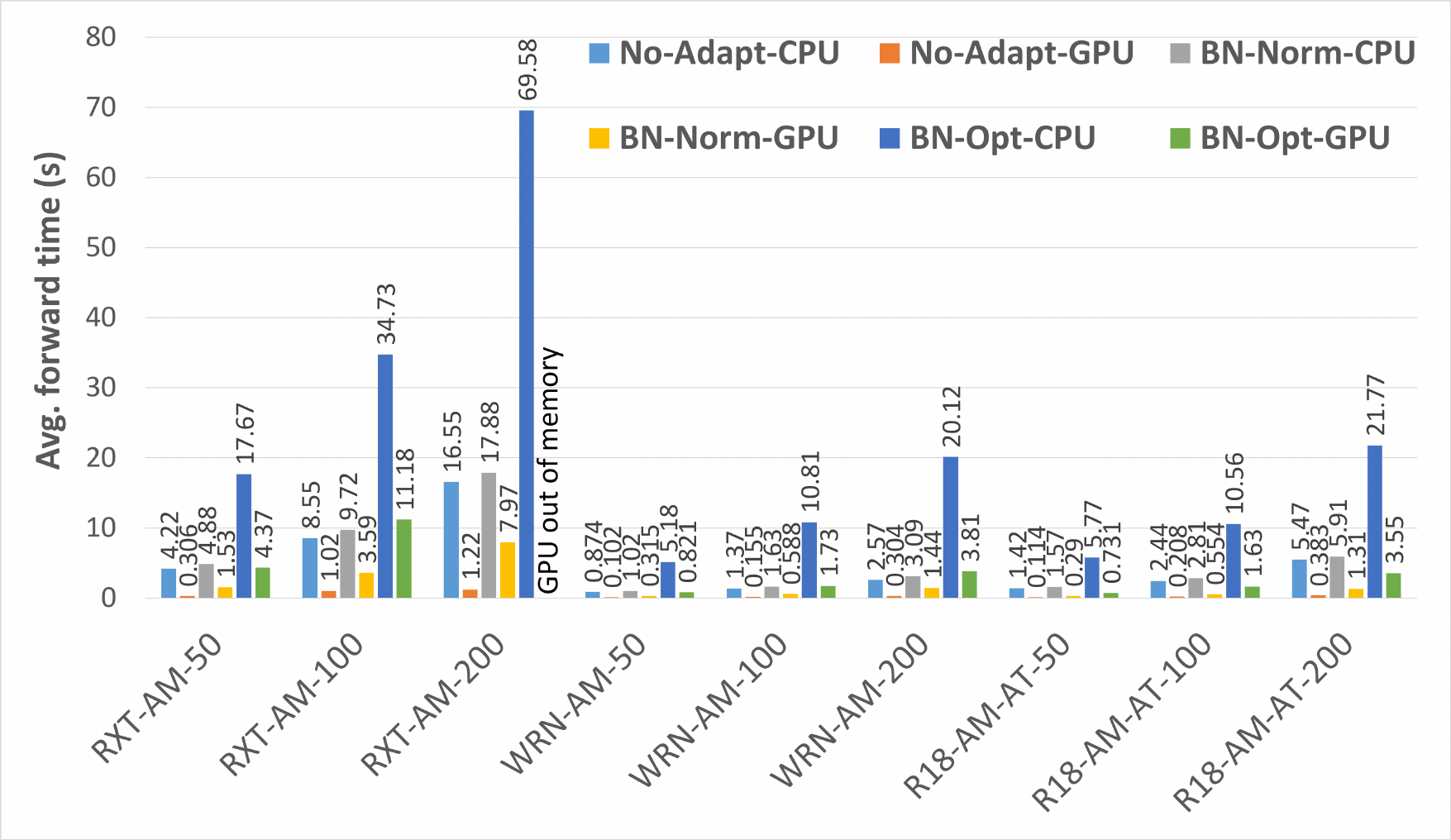}\vspace{-0.0in}
  \caption{Xavier NX forward times (inference + any adaptation)\vspace{-0.2in}}
  \label{fig:nx_adapt}
\end{figure}

\begin{figure*}[t]
        \begin{subfigure}[b]{0.33\textwidth}
                \includegraphics[width=\linewidth,trim=4 4 4 4,clip]{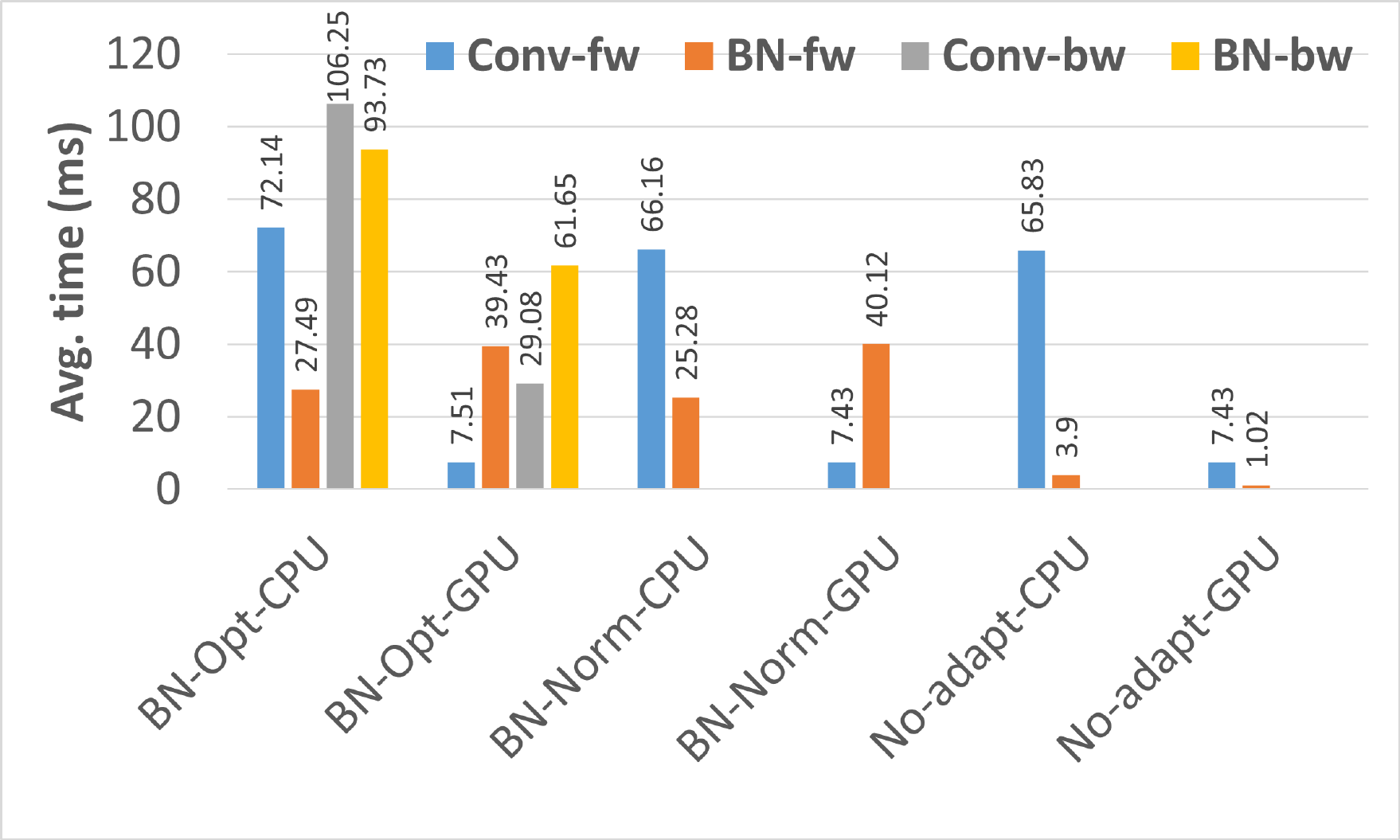}\vspace{-0.0in}
                \caption{ResNeXt\vspace{-0.1in}}
                \label{fig:nx_times_rxt}
        \end{subfigure}%
        \begin{subfigure}[b]{0.33\textwidth}
                \includegraphics[width=\linewidth,trim=4 4 4 4,clip]{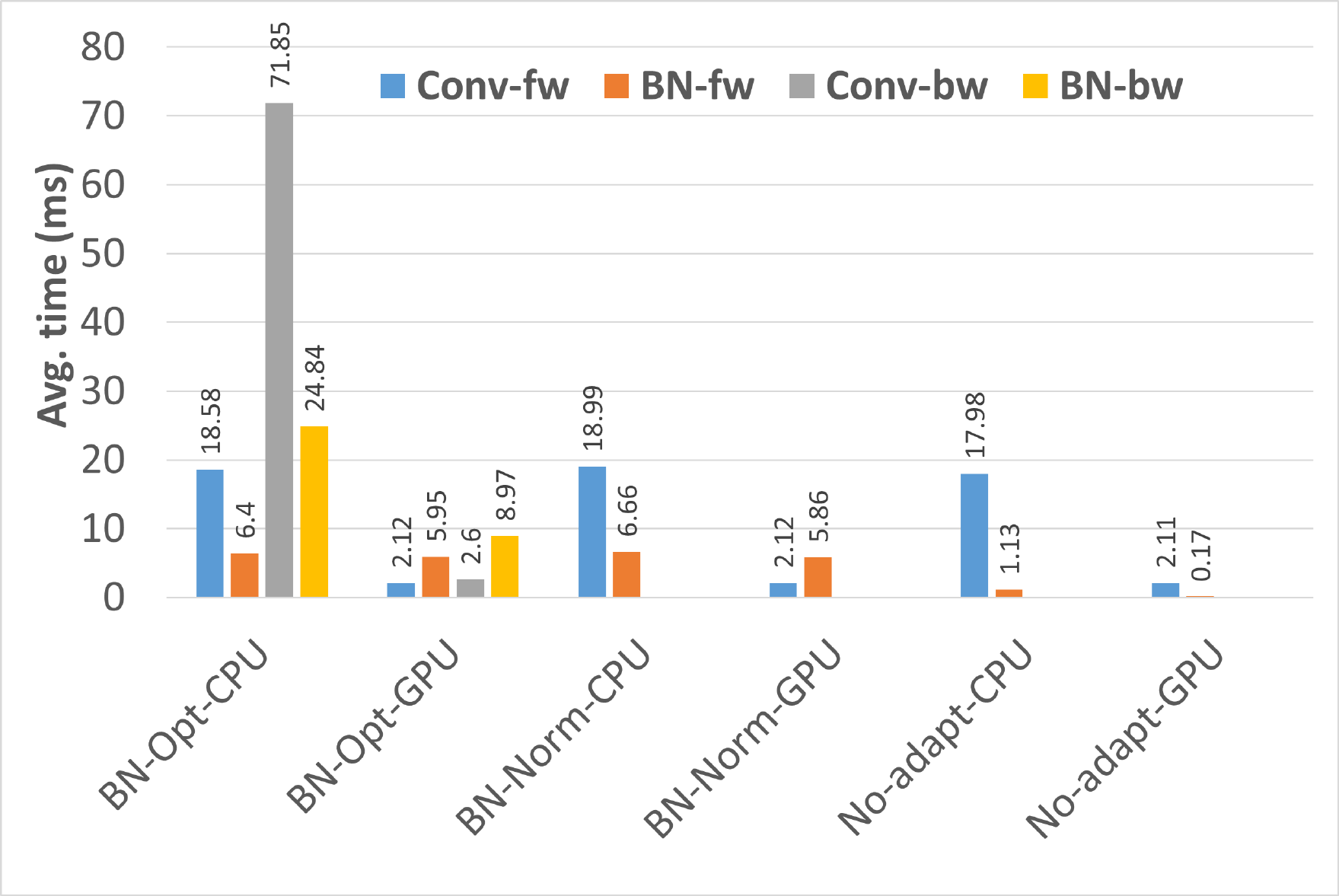}\vspace{-0.0in}
                \caption{Wide-ResNet\vspace{-0.1in}}
                \label{fig:nx_times_wrn}
        \end{subfigure}%
        \begin{subfigure}[b]{0.33\textwidth}
                \includegraphics[width=\linewidth,trim=4 4 4 4,clip]{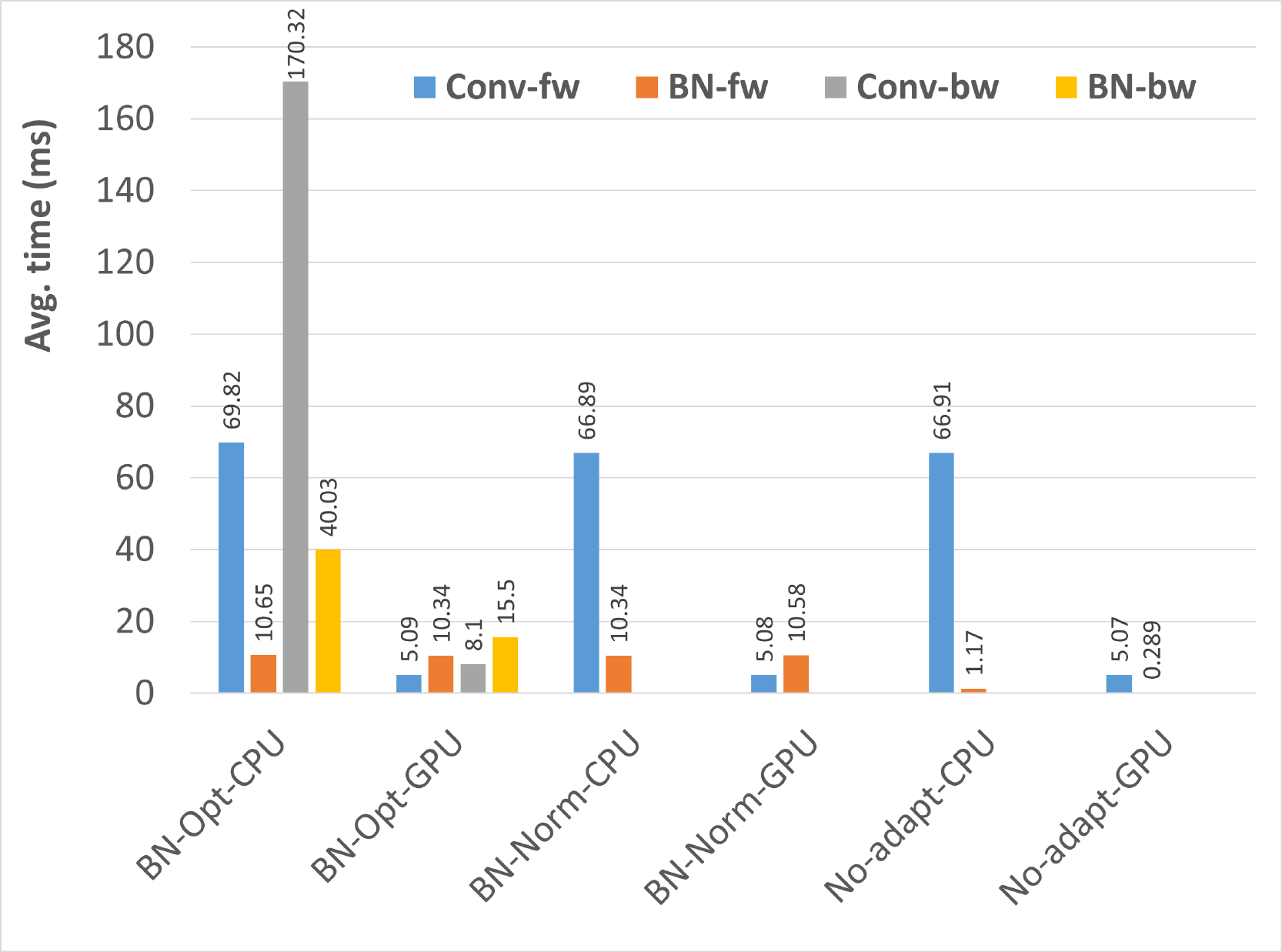}\vspace{-0.0in}
                \caption{ResNet-18\vspace{-0.0in}}
                \label{fig:nx_times_r18}
        \end{subfigure}
        \caption{Xavier NX forward (fw) and any backpropagation backward (bw) time for three DNNs (batch size: 50)\vspace{-0.2in}}\label{fig:nx_breakdown}
\end{figure*}

\vspace{-0.1in}
\subsection{Nvidia Jetson Xavier NX}
\label{subsec:nx}

The forward times of the previous devices are high due to lack of acceleration.
This study evaluates if the adaptation can be accelerated using embedded GPUs (NX has a 384-core Volta). It also includes 6-core Nvidia Carmel Arm, and 8GB memory that is shared between the CPU and the GPU. For CPU results, Arm cores are used to run multi-threaded Pytorch, and for GPU, Pytorch CUDA is used.

{\bf Performance analysis.} Figure~\ref{fig:nx_adapt} shows the performance of No-Adapt, BN-Norm, and BN-Opt for all 9 cases using CPU and GPU. Interestingly, RXT-AM-200 with BN-Opt runs out of memory when executed on the GPU. The dynamic Pytorch graph created for RXT-AM-200 takes up around 5.1GB (when profiled on CPU). But execution on the GPU leads to higher use of memory due to loading of extra cuDNN libraries by Pytorch that are utilized for GPU acceleration. 
For all other cases, speedup using the GPU is observed for the three algorithms and DNN models: for No-Adapt 90.5\% on average, for BN-Norm 68.13\%, and for BN-Opt 79.21\%. 


Figure~\ref{fig:nx_breakdown} shows the average forward and backward times for convolution and BN layers. The convolution backward pass overhead when using GPU for BN-Opt (over forward pass on GPU) is 2.2$\times$ on average. The same difference for CPU on Xavier is 2.5$\times$. Faster forward and backward convolution passes for Xavier explain the speedup obtained for BN-Opt using GPU. 
However, interestingly, the forward BN performance is worse for RXT when using GPU over CPU for both BN-Opt and BN-Norm, but it does not have a major impact on the overall time.



\begin{figure*}[t]
\centering
  \includegraphics[width=0.9\textwidth,trim=4 4 4 4,clip]{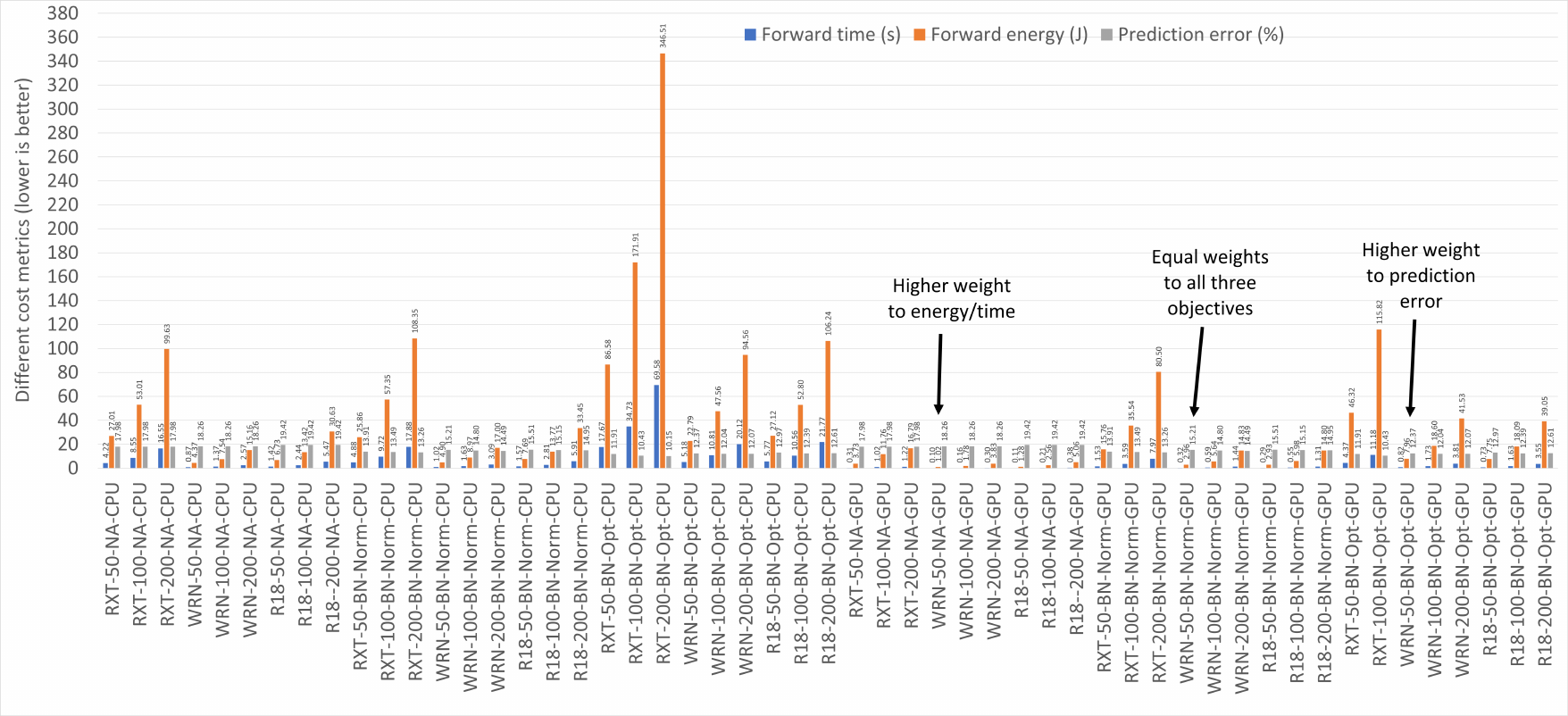}
  \vspace{-0.15in}
  \caption{Performance-energy-accuracy trade-offs analysis for Jetson Xavier NX\vspace{-0.2in}}
  \label{fig:nx_tradeoffs}
\end{figure*}

{\bf Performance-energy-accuracy trade-offs.} Figure~\ref{fig:nx_tradeoffs} shows these cost metrics for the various cases. The important outcomes are: (i) when all three costs are equally weighted, WRN-AM-50 with BN-Norm on GPU leads to overall best solution (0.31 secs, 2.96 J, 15.21\%); (ii) when prediction error has 0.8 weight, WRN-AM-50 with BN-Opt on GPU should be selected (0.82 secs, 7.96 J, 12.37\%). Also, note that the forward time (inference + adaptation) is under 1 sec. While the GPU burns more power than CPU (2.2$\times$), the significantly faster execution of the former makes it more energy-efficient (2.86$\times$); and (iii) if the focus is more on energy or performance, then WRN-AM-50 with No-Adapt (on GPU) is the optimal choice (0.10 secs, 1.02 J, 18.26\%).

{\bf Summary.} Volta in NX leads to significant speedups over CPU for not only No-Adapt and BN-Norm but also for BN-Opt. Interestingly, we see out-of-memory issues with the GPU for RXT. The lightweight Wide-ResNet with BN-Norm algorithm running on GPU is the best when equal priority to given to all three metrics. The same network with BN-Opt (using the GPU) is optimal when accuracy is more important while still giving small weights to performance and energy.  

\vspace{-0.1in}
\subsection{Overall outcomes}
\label{subsec:outcomes}

\begin{figure}[t]
\centering
  \includegraphics[width=0.8\columnwidth,trim=4 4 4 4,clip]{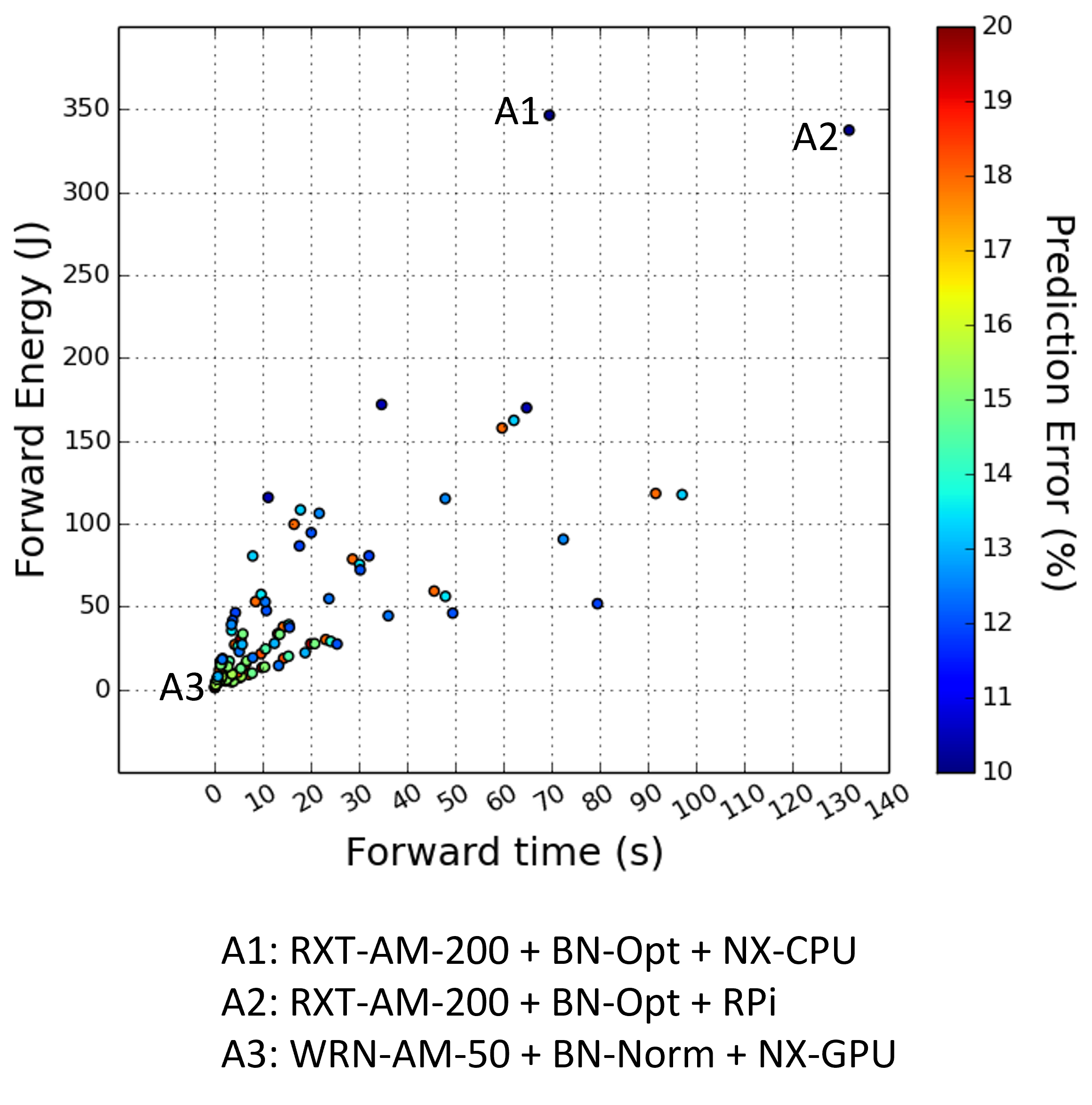}\vspace{-0.1in}
  \caption{Overall results with all the points from Figs.~\ref{fig:pynq_tradeoffs},~\ref{fig:rpi_tradeoffs},~\ref{fig:nx_tradeoffs}. A1/A2: when accuracy is the only priority, A1 shows the lowest runtime and A2 the lowest energy (i.e. among all points with 10.15\% error). A3: optimal point when all three costs are equally important (0.31s, 2.96J, 15.21\%).\vspace{-0.25in}}
  \label{fig:full_dse}
\end{figure}

In this section, we find the overall best DNN model, robust training technique, adaptation approach (with batch sizes), and the device/accelerator.

Figure~\ref{fig:full_dse} shows the complete results, plotting all the design points from Figures~\ref{fig:pynq_tradeoffs},~\ref{fig:rpi_tradeoffs},~\ref{fig:nx_tradeoffs} for the three costs. For an application where prediction accuracy is the only priority, RXT-AM-200 with BN-Opt should be used with the lowest (10.15\%) error. This configuration will require at least 8 GB of memory for execution, therefore Ultra96-v2 is not feasible. Among RPi and NX for this model and adaptation approach, NX with CPU leads to the smallest runtime (A1: 69.58 secs) as GPU runs out of memory for batch size 200. However, in terms of energy, RPi is the most efficient (A2: 337.43 J). But as is evident, both the runtime and energy costs are significant and may not be practical.
{\em Alternatively, if accuracy, performance, and energy are all equally important for an application, then WRN-AM-50 with BN-Norm, running on NX GPU is the best (A3: 0.315 secs, 2.96 J, 15.21\%).} Compared to BN-Opt (on GPU) with WRN-AM-50, BN-Norm outperforms by 61.6\% lower latency and 62.8\% lower energy for the same network. Compared to RXT-AM-200 and BN-Opt that achieved the best accuracy, BN-Norm and WRN-50 on NX GPU is 220$\times$ faster and 114$\times$ more energy-efficient. {\em While WRN-AM-50 with BN-Norm shows 3.05\% lower prediction error than the same network with No-Adapt, the extra overhead for the former is significant for real-time operation: 213 ms and 1.9 J (GPU).}


\subsection{Other edge-based non-robust DNN models}
\label{subsec:others}

We also evaluated MobileNet-V2, a model which is optimized for low-power edge systems. The model size is 9 MB, includes 0.096 GMAC operations, but has 34112 batch-norm parameters (larger than the three robust ResNet models in this work). Closest to MobileNet is WRN (0.33 GMACs, 9 MB, but with 5408 BN parameters). MobileNet is pretrained on CIFAR-10, but without adversarial training, and then evaluated with or without adaptation on CIFAR-10-C. MobileNet showed 81.2\% prediction error without adaptation, which was reduced to 28.1\% using BN-Opt (batch size of 200). This error is still high compared to the three robust models used above (10.15-12.97\% with BN-Opt), highlighting the importance of offline robust training (adversarial training/data augmentation).

Table~\ref{tab:mobile} shows the forward times for MobileNet for various approaches on the NX GPU. For No-Adapt, MobileNet shows, on average, 80.9\%, 19.2\%, 34.8\% better performance than RXT, WRN, and R18, respectively (performance of these models in Figure~\ref{fig:nx_adapt}). However, for both BN-Norm and BN-Opt, it shows an average overhead of $2.1\times$ compared to WRN and R18 as it includes much higher number of BN parameters that are adapted (34112 compared to the two models: 5408, 7808). On the other hand, MobileNet outperforms ResNeXt for BN-Norm and BN-Opt by $2.7\times$ as ResNeXt includes almost $10\times$ more GMACs (1.08) and is closer in terms of number of BN parameters (25216). In summary, while MobileNet is designed for edge, its large number of BN parameters is a bottleneck for BN-Norm and BN-Opt like adaptation algorithms.

\begin{table}
\begin{center}
\begin{tabular}{||c|c|c|c||} 
 \hline
 Batch Size & BN-Opt & BN-Norm & No-Adapt \\ [0.5ex] 
 \hline\hline
 50 & 1.63 s & 0.58 s & 0.07 s \\ 
 \hline
 100 & 3.7 s & 1.18 s & 0.13 s \\
 \hline
 200 & 8.28 s & 2.95 s & 0.25 s \\ [1ex]
 \hline
\end{tabular}
\caption{MobileNet forward time on Xavier NX GPU}\vspace{-0.4in}
\label{tab:mobile}
\end{center}
\end{table}

\vspace{-0.1in}
\subsection{Architecture-algorithm insights}
\label{subsec:insights}

The key algorithm-hardware insights revealed by our study are: {\bf (i)} trade-offs between BN parameters, prediction accuracy, and execution time/memory requirements must be considered when designing a robust DNN for edge. A model with smaller number of BN parameters is more amenable to on-device NN adaptation even though it might not lead to the best improvement in prediction accuracy after adaptation, compared to a model with higher number of BN parameters. For example, Wide-ResNet, with smallest number of BN parameters, is shown to be overall more effective at balancing the various cost metrics during adaptation than ResNeXt and ResNet-18 (and MobileNet);
{\bf (ii)} simply updating BN parameters (BN-Norm) is more suited to test-time adaptation on the edge as it is significantly faster and more energy-efficient than BN-Opt with only 2.45\% increase in prediction error on average. BN-Opt's single backpropagation pass is a major bottleneck. Our results motivate the design of new hardware-aware adaptation algorithms; {\bf (iii)} we show that embedded GPUs are effective in speeding up BN-Norm and BN-Opt. However, the adaptation time is considerable which can lead to real-time constraint violations. For example, even when using BN-Norm with WRN-50 and NX GPU, the extra adaptation time overhead is still significant: 213 ms. As these devices are primarily built for inference, the adaptation overhead can be reduced with custom accelerators designed to support fast BN-based adaptation (and/or backpropagation); 
{\bf (iv)} while the above results are using unpruned and full precision models, pruning and quantization should be explored. However, care must be taken that any model reduction should not compromise the robust accuracy against corruptions; {\bf (v)} online adaptation algorithms are not designed for hardware-constrained edge platforms. Algorithms should minimize memory high water mark (streaming approaches?), and make efficient use of accelerators while minimizing memory overhead of doing so. On the hardware design side, additional MACs and routing fabric would make back propagation less costly, and low power memories including nonvolatile and 3D would enable larger batch sizes with less energy; and {\bf (vi)} as shown by the MobileNet study, online adaptation alone is not sufficient and offline robust training, such as using data augmentation, is also needed to improve accuracy against corruptions.


\vspace{-0.1in}
\section{Related Work}
\label{sec:related}
\vspace{-0.05in}

There has been use of generative adversarial networks (GANs) to design domain-adaptation techniques for DNNs but they do not target unsupervised test-time adaptation~\cite{ganin2015unsupervised, hu2018duplex}. These approaches use a semi-supervised method to improve DNN robustness by adding a domain classifier that discriminates between the source data (used for training) and target data (seen during test time), and tries to maximize the domain classification loss so as to extract features that are domain-invariant. However, these methods require full retraining of the DNN with labeled source data and unlabeled target data, and are therefore not applicable for fast test time adaptation. 

There has been considerable work on characterizing DNN performance on a variety of edge devices, however, they have only targeted inference and not adaptation. Bianco et al. have benchmarked accuracy and inference time of a variety of DNNs on Nvidia Jetson TX1~\cite{bianco2018benchmark}, and found some of the DNNs bottlenecked by the amount of memory available on the device (such as ResNeXt).  Kljucaric et al. characterized the performance of AlexNet and GoogleNet on several devices: Nvidia jetson AGX Xavier, Intel Neural Compute Stick, and Google Edge TPU~\cite{kljucaric2020architectural}. Their results showed best latency for AlexNet is achieved by AGX, while for GoogleNet, TPU is faster. Similar analysis is also targeted for robotics application~\cite{biddulph2018comparing}. An exhaustive characterization of a variety of DNNs on many devices is performed but only for on-edge inference without considering adaptation or robustness~\cite{hadidi2019characterizing}. This analysis included performance, energy, and temperature as well as various pruning/quantization optimizations.

A recent work performed semi-supervised domain adaptation for an internet-of-things edge device (TX2) for gesture recognition and image classification applications~\cite{yang2020mobileda}. A teacher-student technique is used, where knowledge is transferred from a sophisticated teacher model (trained on a server on labeled source data) to a simple student model running on the edge while considering the domain shift in a new environment. However, this method requires initial labeled source data to adapt the student model as well as assumes connection to a server. Hence, it is not meant for unsupervised on-device prediction-time adaptation. Other similar edge-based transfer learning approaches have been introduced for health monitoring at home, but the fine tuning of the deployed model is again performed using supervision and cloud connections~\cite{sufian2021deep}. There are also distributed training based DNN update approaches that do not require connection to a cloud, but they distribute training tasks to multiple connected edge devices. A recent paper proposed a load balancing approach for distributed learning on the edge~\cite{bhattacharjee2020deep}.
\vspace{-0.1in}
\section{Conclusion}
\label{sec:conclusion}
\vspace{-0.05in}

This  paper performed  a  comprehensive measurement study of prediction-time unsupervised DNN adaptation techniques to  quantify their  performance  and  energy  on  various  edge  devices. For each device, the study found the optimal adaptation approach (between BN-Norm and BN-Opt) and the type of robust DNN (among ResNeXt, Wide-ResNet, and ResNet-18) in terms of different cost metrics. A detailed bottleneck analysis of the approaches was also presented, followed by identifying optimization opportunities. Overall, we found that Wide-ResNet with BN-Norm, running on Xavier NX GPU, to be highly effective at balancing the various costs. However, the adaptation overhead is still expensive for meeting tight deadlines. Algorithm-hardware co-design is required to achieve efficient on-device adaptation.

\section*{Acknowledgment}
\addcontentsline{toc}{section}{Acknowledgment}
This work was performed under the auspices of the U.S. Department of Energy by Lawrence Livermore National Laboratory under Contract DE-AC52-07NA27344. IM number: LLNL-CONF-824925.


\bibliographystyle{unsrt}
\bibliography{refs}

\end{document}